\pdfoutput=1

\documentclass[11pt]{article}
\usepackage{CJKutf8}
\usepackage{amssymb}
\usepackage{amsfonts}
\usepackage{booktabs}
\usepackage{authblk}
\usepackage[most]{tcolorbox}
\usepackage{amsmath} 
\usepackage{titlefoot}
\usepackage{pifont} 
\usepackage{marvosym}
\usepackage{authblk}
\usepackage[final]{acl}

\usepackage{times}
\usepackage{latexsym}

\usepackage[T1]{fontenc}

\usepackage[utf8]{inputenc}

\usepackage{microtype}

\usepackage{inconsolata}

\usepackage{graphicx}

\title{Under the Shadow of Babel: How Language Shapes Reasoning in LLMs}

\author{
\bf Chenxi Wang \quad
\bf Yixuan Zhang \quad
\bf Lang Gao \quad
\bf Zixiang Xu \quad \\
\bf Zirui Song \quad
\bf Yanbo Wang \quad
\bf Xiuying Chen\textsuperscript{\Letter} \\
Mohamed bin Zayed University of Artificial Intelligence (MBZUAI) \\
\texttt{\{chenxi.wang, xiuying.chen\}@mbzuai.ac.ae}
}

\begin{document}

\maketitle

\begin{abstract}
Language is not only a tool for communication but also a medium for human cognition and reasoning. If, as linguistic relativity suggests, the structure of language shapes cognitive patterns, then large language models (LLMs) trained on human language may also internalize the habitual logical structures embedded in different languages.
To examine this hypothesis, we introduce BICAUSE, a structured bilingual dataset for causal reasoning, which includes semantically aligned Chinese and English samples in both forward and reversed causal forms. 
Our study reveals three key findings:
(1) LLMs exhibit typologically aligned attention patterns, focusing more on causes and sentence-initial connectives in Chinese, while showing a more balanced distribution in English.
(2) Models internalize language-specific preferences for causal word order and often rigidly apply them to atypical inputs, leading to degraded performance, especially in Chinese.
(3) When causal reasoning succeeds, model representations converge toward semantically aligned abstractions across languages, indicating a shared understanding beyond surface form.
Overall, these results suggest that LLMs not only mimic surface linguistic forms but also internalize the reasoning biases shaped by language. 
Rooted in cognitive linguistic theory, this phenomenon is for the first time empirically verified through structural analysis of model internals.\footnote{\url{https://github.com/Aurora-cx/BabelLLM_public}.}
\end{abstract}

\unmarkedfntext{\Letter: Corresponding Author.}

\section{Introduction}

\begin{figure*}[tb]
    \centering
    \includegraphics[width=1\linewidth]{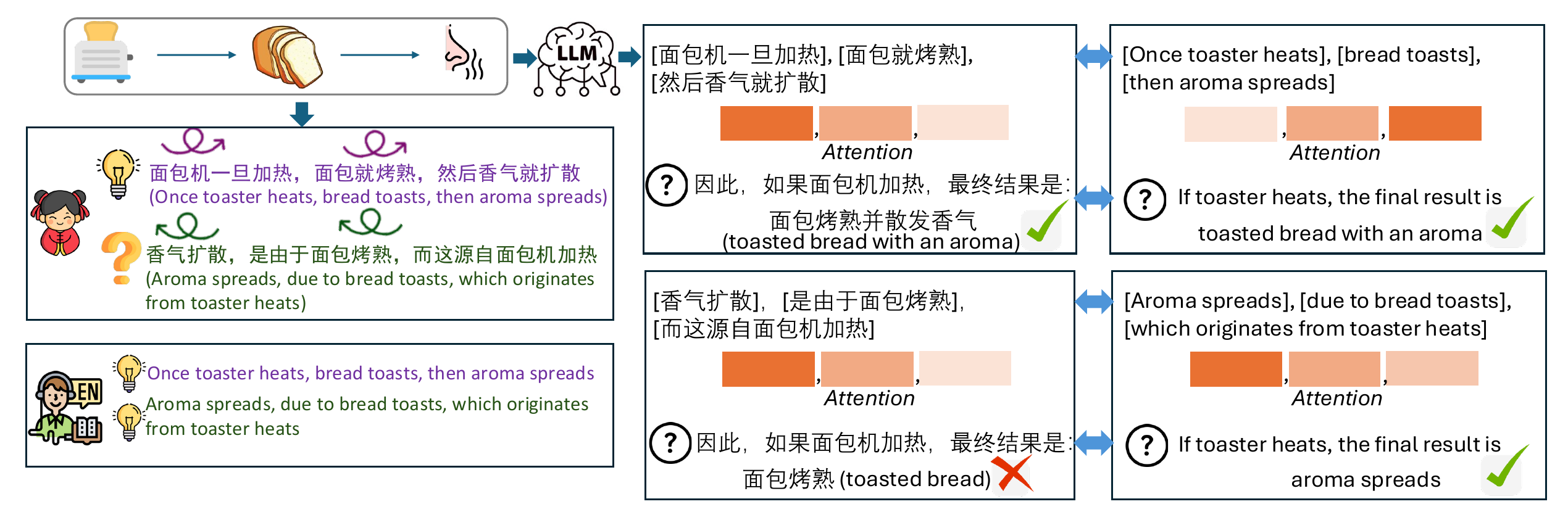}
    \caption{Causal expressions in forward order are common in native Chinese, while reversed forms are rare; both are frequent in English. LLMs internalize such language-specific patterns, leading to divergent reasoning behaviors when processing semantically aligned but structurally different inputs. For Chinese, the model over-applies the prior that the initial phrase signals the cause, misallocating attention to \texttt{[final effect]} and failing to infer correctly. In contrast, English models maintain stable attention and accurate predictions under both orders, revealing the deep influence of syntactic structure on model behavior.}
    \label{fig:main}
\end{figure*}

\emph{“Now the whole world had one language and a common speech… They said, ‘Come, let us build ourselves a city, with a tower that reaches to the heavens…’”  
“The Lord said, ‘…Come, let us go down and confuse their language so they will not understand each other.’”} — Genesis 11:1–7 (NIV)

This is not merely an ancient myth, but a parable of how language both binds and divides us — shaping not only how we speak, but how we think.

This idea lies at the heart of one of the most debated theories in linguistic history—linguistic relativity.  
From Herder and Humboldt in the 18th century to the Sapir–Whorf Hypothesis in the 20th, scholars have long proposed that linguistic structures shape the speaker’s perception of reality~\cite{Humboldt1999-HUMOLO,herder1986origin}.  
Although the hypothesis has been contested, recent empirical work in psycholinguistics—on color terms, spatial reasoning, and temporal concepts—has provided mounting evidence that linguistic variation can influence how humans perceive, infer, and categorize the world.

Given that human reasoning is expressed through language, and that linguistic structures—especially syntactic patterns—constrain how thoughts are formulated, the cognitive models embedded in different languages may also be encoded within language models trained on them.
Modern large language models (LLMs) are primarily trained on internet-scale corpora, which inherently reflect the logical structures and reasoning styles of diverse linguistic communities.
This leads to a critical question: \textbf{Do different languages lead LLMs to think differently? And do LLMs internalize the cognitive biases embedded in different languages?}

This work addresses a critical gap by systematically examining how LLMs represent causal reasoning across languages. We introduce \textsc{BICAUSE}, a structured bilingual dataset of semantically and syntactically aligned causal chains in English and Chinese—prototypical representatives of fusional and analytic language families. The dataset spans multiple domains, tightly controls for lexical and grammatical variation, and includes both sequential and reversed causal chains, enabling interpretable tests of language-specific biases in causal reasoning.

Based on \textsc{BICAUSE}, we develop a fine-grained analytical framework that decomposes each causal chain into 13 interpretable syntactic components and 3 semantic-level causal components. Our analysis proceeds along three axes: 
	1) Syntactic Attention Patterns. We analyze how LLMs distribute attention to individual syntactic components across languages under the same syntactic structure. We observe that LLMs exhibit a greater attentional focus on connectives and subjects when processing Chinese inputs, whereas verbs receive more attention in the English counterparts.
	2) Causal Structure Preferences. We examine the model’s attention distribution over causal components across different language-structure combinations. The results show that LLMs tend to focus more on causal antecedents (causes) in Chinese, while paying greater attention to consequences (effects) in English. Moreover, we find that Chinese-specific causal ordering preferences are internalized and rigidly applied, leading to degraded reasoning accuracy when the input structure deviates from canonical Chinese expressions.
	3) Representation Alignment. Despite the attention divergence, we find that LLMs produce highly similar hidden representations for correctly reasoned samples, regardless of language or syntactic structure. This suggests that divergent attention patterns may converge toward a shared, language-agnostic abstraction space—a finding consistent with recent work on multilingual model alignment~\citep{schut2025think,brinkmann2025latent,anthropic2025graphs}.

Our main contributions are as follows:
	1)	We construct a bilingual Chinese-English causal reasoning dataset \textsc{BICAUSE} designed for fine-grained quantitative analysis. The dataset spans multiple domains and is carefully aligned at both the syntactic and causal component levels, ensuring cross-linguistic comparability and interpretability.
	2)	To the best of our knowledge, this is the first work that systematically analyzes the internal mechanisms of LLMs in cross-lingual causal reasoning from both syntactic and semantic component perspectives. Our approach provides a new lens for understanding how LLMs handle structured reasoning across languages.
	3)	We provide novel empirical evidence that, echoing the claims of linguistic relativity, different linguistic representations not only affect the external reasoning outcomes of LLMs but also lead to the internalization of language-specific causal expression patterns as stable attention allocation strategies within the model.

\section{Related Work}

\paragraph{Linguistic Relativity.} Initially proposed by German scholars Johann Gottfried Herder and Wilhelm von Humboldt, this theory suggests that the structure of a language shapes how its speakers perceive the world~\citep{Humboldt1999-HUMOLO,herder1986origin}.This theory was further developed the well-known Sapir–Whorf Hypothesis~\citep{Whorf1956-WHOLTA,Sapir1929-SAPTSO-2}.While the strong version—that language determines thought—has been widely rejected, the weaker version—that language influences thought—has received substantial empirical support in psycholinguistics and cognitive science~\citep{lucy1992language,brown1954language}.
Studies have shown that speakers of different languages exhibit systematic differences in how they perceive and process concepts such as color~\citep{kay1984sapir}, time~\citep{boroditsky2001language}, and causality~\citep{wolff2009russians}. For instance, the granularity of color terms affects perceptual discrimination, and English and Chinese speakers tend to conceptualize time along horizontal and vertical spatial axes, respectively. Research in bilingualism further reveals that second language acquisition can destabilize and reorganize pre-existing conceptual categories~\citep{l21,l22}.
These findings suggest that language-specific structures encode distinct cognitive schemas~\citep{bisk-etal-2020-experience,piantadosi2022meaning}. We further investigate whether these schemas are internalized by LLMs trained on multilingual corpora.

\paragraph{Cross-Linguistic Performance of LLMs.}
Multilingual LLMs show substantial performance gaps across languages, largely due to the dominance of high-resource languages like English in training data and model design.\citep{xu2025survey, huang2025survey, qin2025survey}. \citet{etxaniz2023think} find that LLMs perform better when prompted in English, and \citet{nie2024decomposed} further show that the syntactic knowledge encoded in these models is also primarily aligned with English. In addition, \citet{chai2022structure} further show that language structure affects cross-lingual transferability, with syntactic composition playing a more central role than word order or co-occurrence. To address these issues, researchers have proposed multilingual benchmarks (e.g., SeaEval \citealp{wang2024seaeval}) and explored approaches such as structural alignment \citep{zhang2024same}, and cross-lingual prompt design \citep{zhang2024autocap} to improve performance in low-resource languages. Building on this, we further explore whether language-specific internal differences in LLMs underlie their cross-lingual performance gaps.

\paragraph{Cross-Lingual Internal Mechanisms in LLMs.}

Recently, researchers have begun to investigate the internal reasoning mechanisms of multilingual language models.~\citet{wendler2024llamas} point out that the abstract “concept space” in models such as LLaMA aligns more closely with English than with other input languages. Building on this,~\citet{schut2025think} show that LLMs tend to make key decisions in representational spaces that are most similar to English, regardless of the input or output language.
Meanwhile, ~\citet{brinkmann2025latent} find that LLMs are capable of encoding shared morphosyntactic concepts across typologically diverse languages.~\citet{wu2024semantic} propose the Semantic Hub Hypothesis, arguing that multilingual meaning is centralized in a shared semantic space and interpreted through the dominant language of the pretraining corpus.~\citet{zhang2024same} further observe that models may invoke the same syntactic circuit—one that is independent of surface language. Anthropic’s recent study also finds that Claude 3.5 Haiku exhibits strong language-agnostic behavior in its middle layers, but its early and output layers remain structurally dominated by English features~\cite{anthropic2025graphs}. However, one important gap remains: whether language-specific patterns of causal expression are internalized by LLMs has not been systematically studied.

\section{Dataset}\label{dataset}

To analyze the causal reasoning behavior of multilingual LLMs, we construct a bilingual dataset named \textbf{BICAUSE} (Bilingual Causal Chains). It covers eight domains, including household routine, natural events, school life, healthcare, shopping and retail, workplace activities, public transportation, leisure and recreation. Each domain contains 50 samples, resulting in 400 samples in total.

\paragraph{Forward Causal Chains.} Each sample in the dataset is a 3-step causal chain, following the structure \texttt{[cause]$\rightarrow$[intermediate effect]$\rightarrow$[final effect]}, where \texttt{cause} leads to \texttt{intermediate effect}, and \texttt{intermediate effect} leads to \texttt{final effect}. The English and Chinese versions adopt parallel syntactic patterns, where all Chinese phrases are semantically equivalent to their English counterparts:

\begin{tcolorbox}[colback=gray!10, left=0mm, right=0mm, top=1mm, bottom=1mm] 
\small  
\texttt{[Once][subject$_1$][verb$_1$],}
\texttt{[subject$_2$][verb$_2$],}

\texttt{[then][subject$_3$][verb$_3$].}
\end{tcolorbox}

\begin{tcolorbox}[colback=gray!10, left=0mm, right=0mm, top=1mm, bottom=1mm] 
\small  
\texttt{[\begin{CJK*}{UTF8}{gbsn}一旦
\end{CJK*}][subject$_1$][verb$_1$],}
\texttt{[subject$_2$][verb$_2$],}

\texttt{[\begin{CJK*}{UTF8}{gbsn}然后
\end{CJK*}][subject$_3$][verb$_3$].}
\end{tcolorbox}

Here is a specific example:
\begin{tcolorbox}[colback=gray!10, left=0mm, right=0mm, top=1mm, bottom=1mm] 
\small  
Once toaster heats, bread toasts, then aroma spreads. 

\begin{CJK*}{UTF8}{gbsn}  
一旦面包机加热，面包就烤熟，然后香气就扩散。
\end{CJK*}
\end{tcolorbox}

We also provide a QA-style inference task:
\begin{tcolorbox}[colback=gray!10, left=0mm, right=0mm, top=1mm, bottom=1mm] 
\small  
[Question-en]: Therefore, if toaster heats, the final result is

[ Answer-en ]: Aroma spreads.

[Question-zh]:\begin{CJK*}{UTF8}{gbsn}  
因此，如果面包机加热，最终结果是
\end{CJK*}

[ Answer-zh ]:\begin{CJK*}{UTF8}{gbsn}  
香气扩散。
\end{CJK*}
\end{tcolorbox}
Each sample is annotated with conjunction components and three causal components, each consisting of a labeled \texttt{[subject,verb]} pair. The question part in the QA format is also decomposable.

\paragraph{Reversed Causal Chains.}We also construct reversed causal chains using the structure \texttt{[final effect]$\rightarrow$[intermediate effect]$\rightarrow$[cause]}. The following examples show the structure of reversed causal chains in English and Chinese.

\begin{tcolorbox}[colback=gray!10, left=0mm, right=0mm, top=1mm, bottom=1mm] 
\small  
\texttt{[subject$_3$][verb$_3$],}
\texttt{[due to][subject$_2$][verb$_2$],}

\texttt{[which originates from][subject$_1$][verb$_1$].}
\end{tcolorbox}

\begin{tcolorbox}[colback=gray!10, left=0mm, right=0mm, top=1mm, bottom=1mm] 
\small  
\texttt{[subject$_3$][verb$_3$],}
\texttt{[\begin{CJK*}{UTF8}{gbsn}是由于\end{CJK*}][subject$_2$][verb$_2$],}

\texttt{[\begin{CJK*}{UTF8}{gbsn}而这源自\end{CJK*}][subject$_1$][verb$_1$].}
\end{tcolorbox}

Here is a specific example:

\begin{tcolorbox}[colback=gray!10, left=0mm, right=0mm, top=1mm, bottom=1mm] 
\small  
Aroma spreads, due to bread toasts, which originates from toaster heats.  

\begin{CJK*}{UTF8}{gbsn}  
香气扩散，是由于面包烤熟，而这源自面包机加热。
\end{CJK*}
\end{tcolorbox}

\paragraph{Dataset Characteristics.}
We evaluate BICAUSE on Qwen1.5-1.8B-Chat~\citep{qwen}. The model achieves balanced accuracy across English and Chinese, with average scores of 91\% and 91.2\%, respectively. Detailed results for each domain and comparisons with other models are reported in Appendix~\ref{accuracy}.

BICAUSE is designed with the following key properties:
1) Domain diversity. It covers 8 realistic topics to ensure rich reasoning contexts.
2) Semantic alignment. Every component is strictly aligned between English and Chinese.
3) Syntactic consistency. All components follow a standardized format and consistent connector structures.
4) Modular causal components. Each chain explicitly contains 3 causal components.

In sum, \textsc{BICAUSE} offers a high-quality, cross-lingual benchmark for interpretable causal reasoning. Its semantic alignment and modular design make it well-suited for analyzing how LLMs internalize causal structures across different languages.

\paragraph{Model Selection.} 
We primarily focus our analysis on Qwen1.5-1.8B-Chat~\citep{qwen}, as it serves as an ideal “toy model” for studying multilingual reasoning in LLMs. Its architecture remains relatively close to a vanilla Transformer, with moderate complexity—24 layers and 16 attention heads per layer—making it more interpretable than larger or heavily optimized models.
In addition, it demonstrates strong performance in multilingual understanding \cite{qwen2024report}, reducing the risk of English-centric bias and ensuring that the insights derived from both English and Chinese prompts are equally robust in terms of reliability.
We further explore our findings on models from different families and with larger parameter sizes, including LLaMA3.2 (1B/3B), Mistral-7B and Qwen1.5 (32B/72B), and observe that the conclusion that models internalize language-specific expression patterns is generalizable (see Appendix~\ref{accuracy},~\ref{appendix_causal} for further details).

\section{Cross-Lingual Attention Patterns over Syntactic Components}

To investigate whether language-specific syntactic preferences are internalized by LLMs, we focus on analyzing attention patterns. Since attention directly governs how a token "sees" others during inference, it provides a window into how the model structurally organizes information and encodes grammatical relationships.

In this section, we decompose each Chinese and English causal chain into 13 interpretable syntactic components and quantitatively compare attention allocation patterns across languages. 
Our analysis reveals that while the layerwise attention trajectories over corresponding components are highly consistent between Chinese and English, the absolute attention weights exhibit clear divergences, suggesting that LLMs internalize language-specific structural biases and expression preferences.
\subsection{Analytical Framework}

To investigate how LLMs allocate attention across syntactic roles in cross-lingual causal reasoning, we begin with structurally aligned sequential causal chains in Chinese and English (e.g., [cause] $\rightarrow$ [intermediate effect] $\rightarrow$ [final effect]). This natural ordering in time and logic provides a controlled setting to compare attention allocation patterns.

We segment each causal chain into 13 interpretable syntactic components, including three subject-verb pairs for the cause, intermediate, and final effect (e.g., ``toaster–heats,'' ``bread–toasts,'' ``aroma–spreads''), the subject-verb pair in the question sentence, four frequent causal connectives (``once,'' ``then,'' ``if,'' ``therefore''), and the “final result” trigger word itself. The Chinese versions are manually aligned with their English counterparts to ensure cross-linguistic comparability.

\paragraph{Measuring Component-Level Attention.}
We define a general metric to quantify the attention received by any component (syntactic or semantic) at each layer.
This formulation allows us to analyze both syntactic components and causal components under a unified attention-based framework. 

Let $A^{(l,h)} \in \mathbb{R}^{T \times T}$ denote the attention matrix from layer $l$ and head $h$ with sequence length $T$. For each token index $i$ in the component token set $\mathcal{T}_c$, we consider only the \emph{valid queries} $j > i$ due to the autoregressive attention mask.
We compute, for each component $c$, the attention ratio per layer and head as:
\begin{equation}
    r^{(l,h)}_c = \frac{1}{|\mathcal{T}_c|} \sum_{i \in \mathcal{T}_c} \frac{\sum_{j = i+1}^{T} A^{(l,h)}_{j,i}}{\sum_{j = i+1}^{T} \sum_{k=1}^{T} A^{(l,h)}_{j,k}}.
\end{equation}

This yields a $L \times H$ matrix per component, where $L$ is the number of layers and $H$ is the number of heads.
We define the \textbf{Relative Component Attention Ratio} (RCAR) at layer $l$ as:
\begin{equation}
    \mathrm{RCAR}_c^{(l)} = \sum_{h=1}^{H} r^{(l,h)}_c.
\end{equation}

To ensure fair cross-lingual comparisons, we apply a two-level normalization strategy. First, we compute the proportion of attention received by each token from all valid queries (i.e., subsequent positions in autoregressive decoding), which controls for differences in sequence length and visibility range. Second, we average attention scores across tokens within a component to neutralize the effect of token count on total attention mass. This allows us to compare attention trends across languages and layers while mitigating biases.

\subsection{Cross-Lingual Similarities}

\begin{figure*}[tb]
    \centering
    \includegraphics[width=1\linewidth]{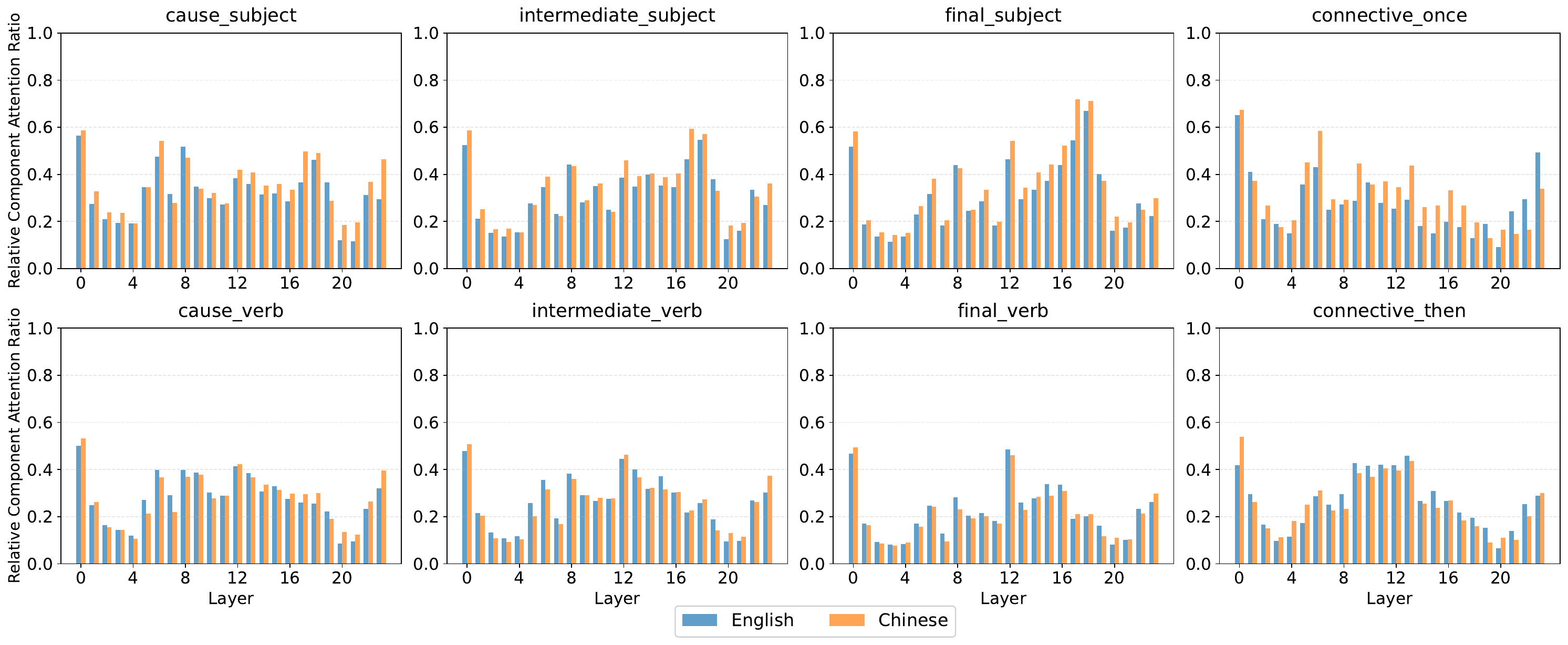}
    \caption{Layerwise RCAR over eight syntactic components in Chinese (orange) and English (blue) causal chains. Despite similar layerwise trends, Chinese shows higher attention on subjects, while English focuses more on verbs.}
    \label{fig:syntax1}
\end{figure*}
Despite the fact that Chinese and English belong to analytic and fusional language families respectively, we observe striking consistency in how LLMs allocate attention across syntactic components over different layers during causal reasoning. As shown in Figure~\ref{fig:syntax1}, the attention trajectories of 8 core components (see Appendix~\ref{qwen_all} for others) exhibit remarkably similar trends across languages.

Specifically, attention to subjects is consistently higher than to verbs in both Chinese and English, and \texttt{[once]} receives more attention than \texttt{then} on average. Attention to the \texttt{[cause subject]} is relatively evenly distributed across layers, while \texttt{[final subject]} shows a prominent peak at layers 17–18. The patterns for \texttt{[cause verb]} and \texttt{[intermediate verb]} are similar, whereas \texttt{[final verb]} receives noticeably less attention. These shared layerwise attention dynamics may reflect structural attention preferences that the model has naturally acquired during pretraining, shaped by the characteristics of its training corpus and learning objective~\citep{tenney2019bert,clark2019does,voita2020analyzing}.

Such consistency is also attributable to the structured nature of our dataset—where Chinese and English samples are tightly aligned at both semantic and syntactic levels—as well as the multilingual pretraining of the LLM, which encourages the development of language-agnostic representations. Moreover, syntactic components such as subjects, verbs, and connectives often fulfill analogous discourse functions across languages. Even when their surface realizations differ, the model learns to allocate attention in consistent patterns based on shared functional roles. This finding resonates with prior work suggesting that LLMs achieve conceptual alignment across languages through a unified representational space~\citep{schut2025think,brinkmann2025latent,anthropic2025graphs}.

\subsection{Typological Divergences}

\begin{figure}[t]
  \includegraphics[width=0.9\columnwidth]{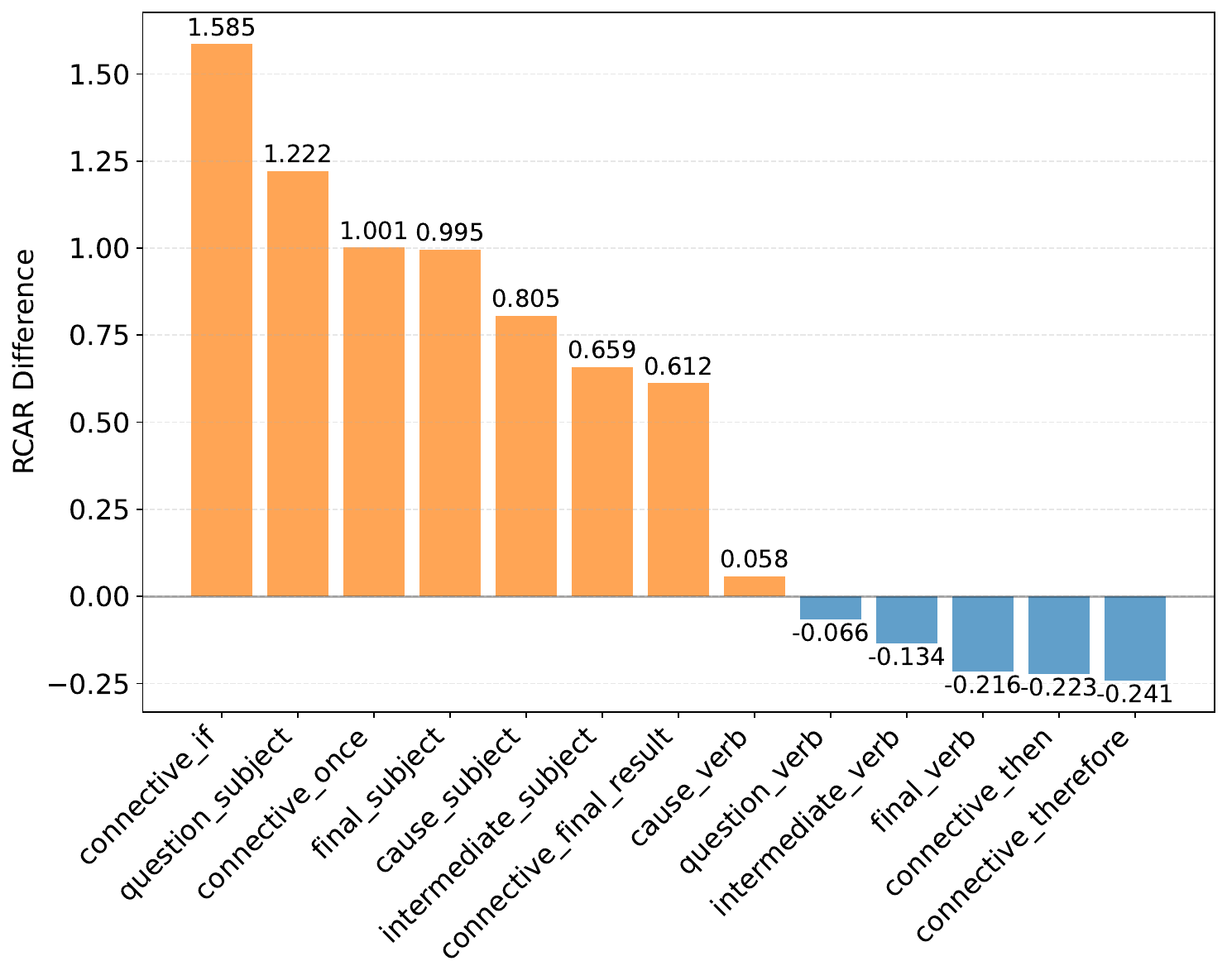}
  \caption{RCAR differences between Chinese and English across 13 syntactic components. Each bar represents the difference in the sum of RCARs assigned to a given component across all layers, computed as (Chinese - English).}
  \label{fig:component_diff_bar}
\end{figure}

Despite the strong structural alignment observed in the layerwise attention trajectories across languages, we find that the RCARs assigned to specific syntactic components differ systematically.
Figure~\ref{fig:component_diff_bar} presents the component-wise attention differences between Chinese and English. Clear disparities emerge in attention preferences: for Chinese inputs, the model tends to focus more on subjects and conditional connectives such as "once" and "if"; in contrast, English inputs elicit higher attention to verbs and logical progression connectives like "then" and "therefore."

These differences can be partially attributed to typological factors. Chinese is classified as a topic-prominent language \citep{li1976subject}, where discourse structure emphasizes participants and global context. This aligns with the model’s higher attention toward subjects and sentence-initial causal markers in Chinese. English, on the other hand, is considered a verb-centric and tense-driven language \citep{halliday1994functional}, with verb morphology carrying core semantic content. As a result, the model allocates more attention to verbs and result-oriented connectives in English inputs. 
Additionally, we observe that causal connectives such as "once" and "if" in Chinese tend to occur at the beginning of a sentence, guiding the conditional structure, whereas “then” and “therefore” in English often appear mid-sentence or post-verbally to indicate reasoning progression. This structural difference causes the model to adjust its attention focus between languages accordingly.

It is also worth noting that Chinese is a pro-drop language, where subjects are frequently omitted, especially when the referent is contextually accessible~\citep{pro1,pro2,pro3}. Consequently, when subjects are explicitly stated, they often carry greater informational weight—potentially contributing to the model’s increased attention toward subject components in Chinese.

Typology thus offers a compelling explanation: these attention disparities likely reflect statistical distributional differences in pretraining corpora, rooted in the distinct syntactic conventions of expressing causality across languages. Our findings suggest that even in tightly aligned causal chains, multilingual LLMs adjust their internal attention mechanisms in ways that mirror the discourse logic of each input language.

\section{Cross-Lingual Attention Patterns over Causal Components}

\begin{figure*}[tb]
    \centering
    \includegraphics[width=0.8\linewidth]{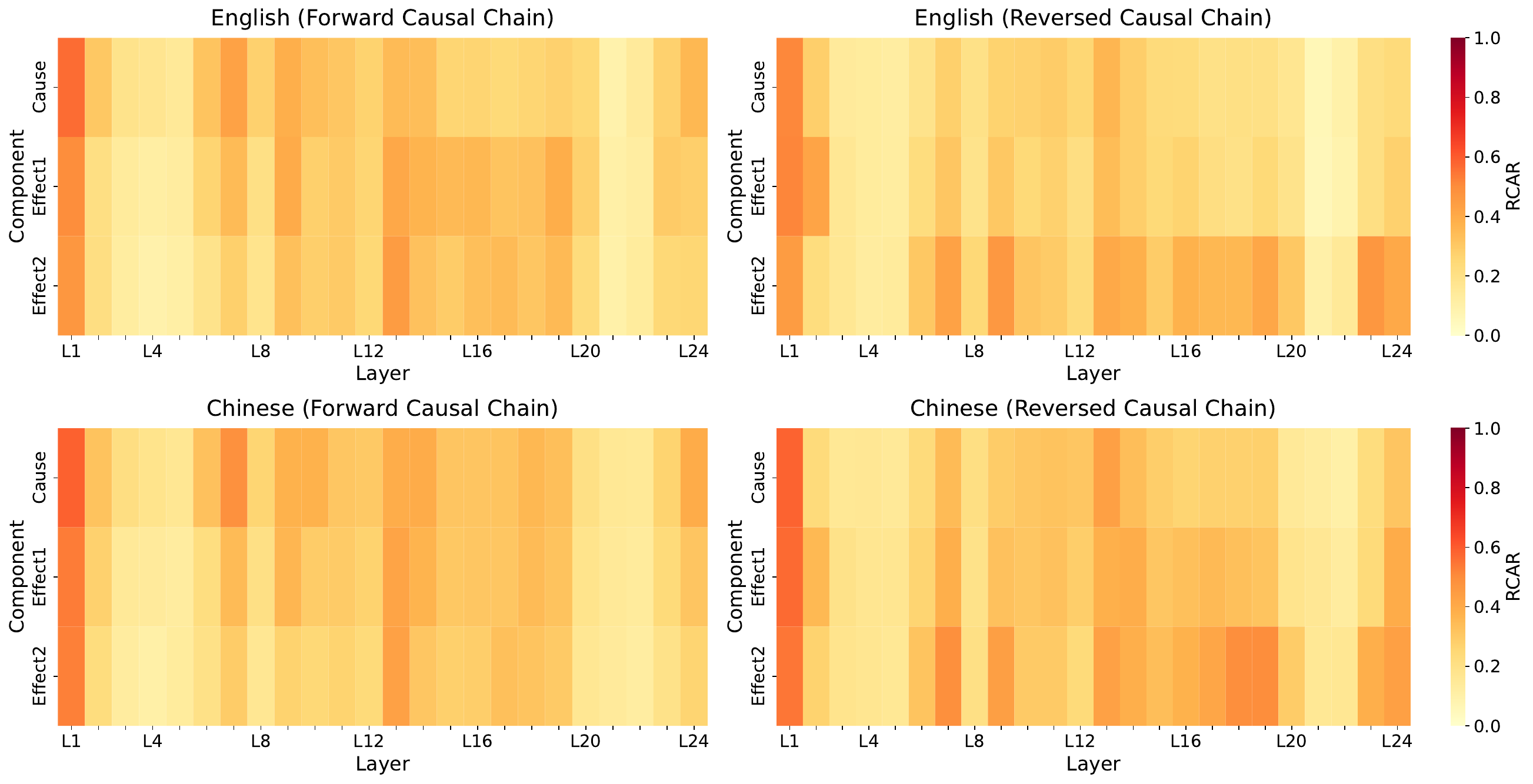}
    \caption{LLayerwise RCAR heatmaps for three core causal components — \texttt{[cause]}, \texttt{[intermediate effect]}, and \texttt{[final effect]} — across four input conditions: \textbf{English (Forward Causal Chain)} (top left), \textbf{English (Reversed Causal Chain)} (top right), \textbf{Chinese (Forward Causal Chain)} (bottom left), and \textbf{Chinese (Reversed Causal Chain)} (bottom right). Warmer colors indicate higher attention allocation.}
    \label{fig:heatmap}
\end{figure*}

In this section, we further examine how multilingual LLMs allocate attention to different causal components across languages and syntactic structures. Building on the component segmentation introduced in Section~\ref{dataset}, we compute the RCAR for each of the three causal components by layer.
We additionally introduce the Singular Vector Canonical Correlation Analysis (SVCCA)~\citep{raghu2017svcca}, a tool for quickly comparing two representations, to measure the overall similarity of layerwise attention patterns across different sentence structures and lanuages.

Our findings reveal striking differences in attentional focus between Chinese and English, even when the underlying causal chains are semantically and syntactically aligned. Moreover, within Chinese, we observe a rigid transfer of the model’s internalized causal ordering preference when confronted with reversed syntactic structures—leading to a drop in reasoning accuracy. In contrast, the English attention distribution appears more adaptive to structural changes, reflecting greater flexibility in its learned causal reasoning schema.

\subsection{Structural Sensitivity Measured by SVCCA}

To assess the consistency of attention structures across different sentence forms, we compute SVCCA similarity scores between the 24-layer attention distributions of four input variants: \textit{English forward causal chains}, \textit{Chinese forward causal chains}, \textit{English reversed causal chains}, and \textit{Chinese reversed causal chains}. Specifically, we extract a 3-dimensional vector from each layer representing the RCAR scores of the three causal components: \texttt{[cause]}, \texttt{[intermediate effect]}, and \texttt{[final effect]}. This results in a $24 \times 3$ attention trajectory matrix for each input type. SVCCA is then used to quantify the layerwise similarity between different structures.

The results reveal that the original English and Chinese chains exhibit a high SVCCA similarity of 0.73. In contrast, the similarity between the English forward and reversed chains drops to 0.64. Moreover, comparisons between the two Chinese input forms show low scores around 0.46.
Interestingly, these SVCCA similarity patterns closely mirror the model’s reasoning accuracy. The English and Chinese forward chains achieve comparable performance (91.0\% and 91.2\%, respectively), consistent with their SVCCA score. The English reversed chains yield a lower accuracy of 88.5\%, while the Chinese reversed chains perform the worst at 76.5\%, mirroring the decreasing SVCCA score. This trend suggests that the layerwise evolution of attention is not merely a structural byproduct, but instead encodes functional representations that are critical to causal reasoning.

\subsection{Cross-Linguistic and Structural Divergences}

\paragraph{Forward Causal Chains.}
Figure~\ref{fig:heatmap} presents the RCAR-based attention distributions over causal components across all layers for four input types. In forward causal chains, Chinese inputs exhibit a clear “cause $\rightarrow$ effect” preference—models allocate the highest attention to the \texttt{[cause]} component, followed by \texttt{[intermediate effect]} and \texttt{[final effect]}. This preference aligns with findings in linguistic typology. Chinese is a topic-prominent language that emphasizes event continuity and often relies on preposed topics to maintain discourse coherence. Its sentence structure reflects a traditional\begin{CJK*}{UTF8}{gbsn}  
“起-承-转-合”
\end{CJK*}(\textit{“introduction–development–transition–conclusion”}) pattern, favoring linear unfolding of causal chains \textit{starting from the event origin.}

In contrast, English models show a more evenly distributed attention pattern, with downstream components receiving even higher attention than the cause. This flexibility reflects the syntactic diversity of English, which lacks a strong preference for forward causal order. Causal expressions in English frequently adopt result-first constructions, such as “X happened, because Y”.

\paragraph{Cross-Linguistic Narrative Preferences.}
Chinese and English exhibit notable preferences in causal sequencing. Chinese heavily relies on forward causal structures, whereas English, being more linear and focus-oriented, often presents information in a “focus-first” manner. When causal chains are expressed in reversed order, the behavioral gap between models becomes more apparent. 

In paraphrased Chinese inputs, the model disproportionately focuses on the sentence-initial \texttt{[final effect]} instead of the true causal origin (\texttt{[cause]}), indicating an overreliance on the prior that “sentence-initial = cause.” When the cause is embedded in a relative clause, its topic accessibility diminishes, making it harder for the model to trace the causal core. While reversed causal structures are grammatically valid in Chinese, they are rare in native usage and likely underrepresented in the training corpus, weakening the model’s ability to construct semantic paths for such structures—ultimately leading to lower accuracy. In contrast, reversed structures are prevalent in English corpora, enabling the model to learn their patterns more effectively.

This attention bias hinders the model’s ability to correctly capture the causal logic, explaining the performance drop observed in paraphrased Chinese structures.
By comparison, English paraphrases have a milder impact. Since reversed causal expressions are common in English training data, the model demonstrates greater generalizability and is able to extract causal information robustly across structures.

These findings suggest that even when expressing the same semantic content, structural and pragmatic differences between languages significantly affect LLMs’ internal representations and reasoning paths. This highlights the importance of accounting for language-specific structural preferences in cross-lingual modeling, rather than focusing solely on semantic alignment.

\section{Representation Similarity Across Languages and Structures}

\begin{figure}[t]
  \includegraphics[width=\columnwidth]{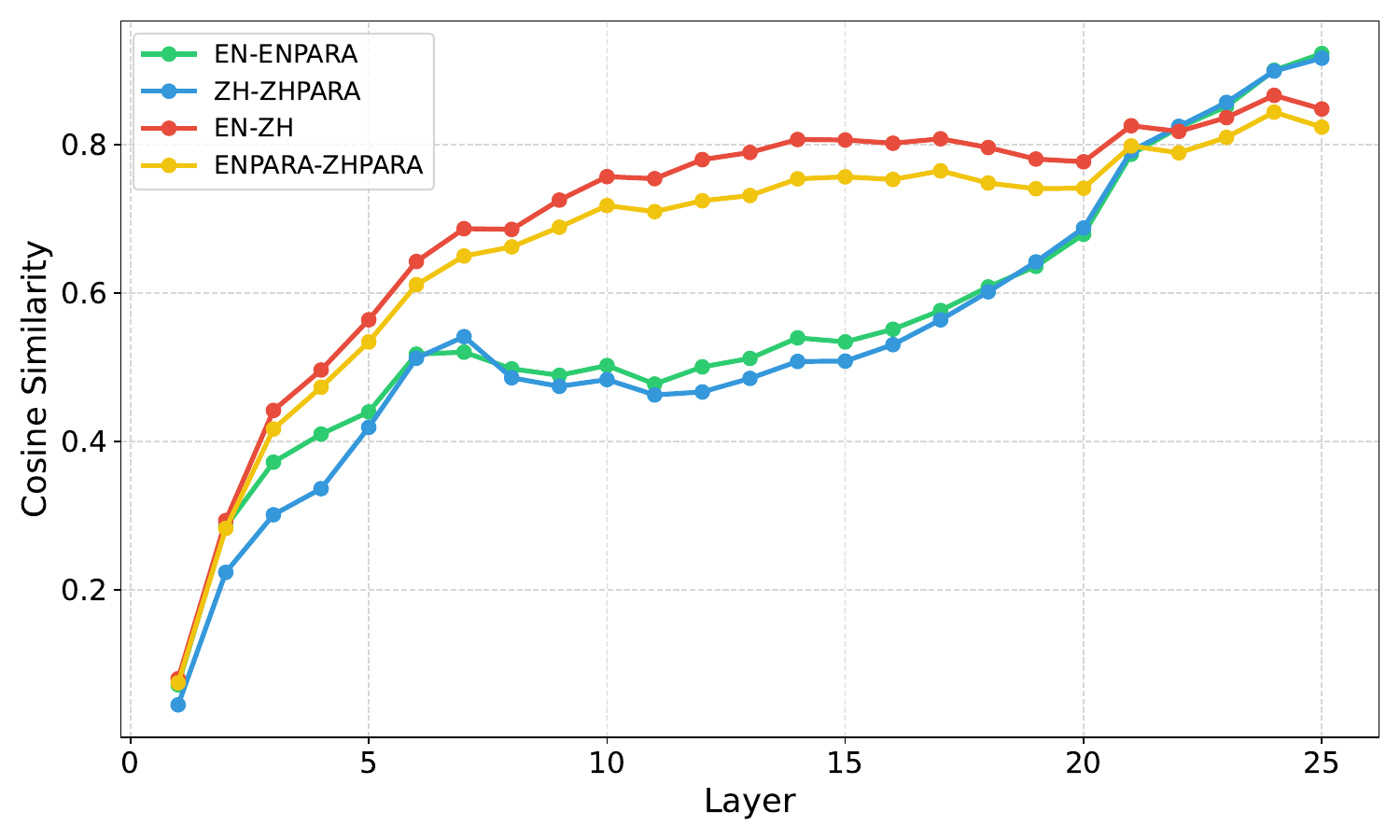}
  \caption{Layerwise cosine similarity of hidden vectors at the final causal token position between input pairs where both predictions are correct. Each line represents a different pairwise comparison.
}

  \label{fig:causal_hidden_cosine}
\end{figure}

We further analyze the differences in representations at the convergence point of causal information. Specifically, for each input sample, we extract the hidden vector at the final token of the causal chain—this position represents the model’s state after integrating the causal information~\cite{meng2022locating,geva2023dissecting,yang2024large}. For each pair of input forms (e.g., English vs. Chinese), we compute the layerwise cosine similarity between the hidden vectors at this position across all layers, considering only samples where both inputs produce correct predictions.

As shown in Figure~\ref{fig:causal_hidden_cosine}, the English–Chinese pair exhibits the highest similarity across most layers, indicating that when semantics and structure are aligned, the model’s internal reasoning paths are highly synchronized. In contrast, the Chinese–reversed Chinese pair shows the lowest similarity in the middle layers, suggesting that word order changes disrupt the internal representation more strongly in Chinese. This aligns with our earlier findings.
Notably, all input forms converge to relatively high cosine similarity in the final layers, implying that once the model succeeds in completing the reasoning task, it forms functionally equivalent causal representations, regardless of surface structure. This phenomenon also echoes prior works~\citep{schut2025think,anthropic2025graphs}.

\section{Conclusion}
This paper introduces BICAUSE, a structured bilingual dataset designed to support fine-grained analysis of LLMs’ causal reasoning. Our findings include:
	1.	Even when semantics and structure are fully aligned, LLMs exhibit typologically aligned attention patterns;
	2.	LLMs internalize language-specific habitual causal constructions;
	3.	When reasoning succeeds, models form a shared understanding that goes beyond surface form.
    
These results reveal the language-specific biases of LLMs in causal reasoning and their impact on performance.
Future work includes extending BICAUSE and advancing research on multilingual interpretability and alignment mechanisms.

\section*{Limitations}
This study focuses exclusively on Chinese and English. While these two languages differ significantly in typology, they represent only a small portion of the world’s linguistic diversity. We selected this language pair to ensure high-quality semantic alignment, as the authors are fluent in both languages and could manually verify cross-linguistic equivalence. Future work may extend this research to a broader set of languages, especially those with greater structural divergence or lower resource availability.
In addition, although we analyze eight models, our coverage does not encompass all architectures or training paradigms.
This study primarily focuses on attention patterns and hidden representations; future research could incorporate additional interpretability methods to provide a more comprehensive analysis.

\bibliography{custom}

\appendix

\section{Accuracy Results on BICAUSE}
\label{accuracy}
This section presents the performance of all models mentioned on the BICAUSE dataset. See Table~\ref{tab:model_accuracy_by_language} and Table~\ref{tab:model_accuracy_by_para}.

\begin{table*}[htbp]
\caption{Accuracy (\%) for English and Chinese across different models and domains. Domain abbreviations: House=household routine, Nature=natural events, School=school life, Health=healthcare, Shop=shopping retail, Work=workplace activities, Trans=public transportation, Leisure=leisure recreation.}

\centering
\small
\begin{tabular}{lccccccccc}
\toprule
\textbf{Model} & \textbf{House} & \textbf{Nature} & \textbf{School} & \textbf{Health} & \textbf{Shop} & \textbf{Work} & \textbf{Trans} & \textbf{Leisure} & \textbf{Avg} \\
\midrule
Llama-3.2-1B (En) & 52.0 & 42.0 & 54.0 & 72.0 & 72.0 & 74.0 & 64.0 & 58.0 & 61.0 \\
Llama-3.2-1B (Zh) & 44.0 & 48.0 & 40.0 & 66.0 & 44.0 & 42.0 & 64.0 & 50.0 & 49.8 \\
Llama-3.2-3B (En) & 36.0 & 42.0 & 24.0 & 38.0 & 6.0 & 14.0 & 40.0 & 16.0 & 27.0 \\
Llama-3.2-3B (Zh) & 60.0 & 64.0 & 50.0 & 44.0 & 62.0 & 42.0 & 78.0 & 54.0 & 56.8 \\
Mistral-7B (En) & 94.0 & 98.0 & 98.0 & 96.0 & 94.0 & 94.0 & 94.0 & 98.0 & 95.8 \\
Mistral-7B (Zh) & 86.0 & 86.0 & 82.0 & 88.0 & 92.0 & 92.0 & 94.0 & 88.0 & 88.5 \\
Qwen1.5-1.8B (En) & 84.0 & 84.0 & 96.0 & 92.0 & 88.0 & 96.0 & 100.0 & 88.0 & 91.0 \\
Qwen1.5-1.8B (Zh) & 90.0 & 90.0 & 94.0 & 94.0 & 94.0 & 82.0 & 96.0 & 90.0 & 91.2 \\
Qwen1.5-14B (En) & 100.0 & 98.0 & 98.0 & 98.0 & 94.0 & 94.0 & 98.0 & 94.0 & 96.8 \\
Qwen1.5-14B (Zh) & 94.0 & 98.0 & 98.0 & 96.0 & 98.0 & 94.0 & 96.0 & 94.0 & 96.0 \\
Qwen1.5-32B (En) & 100.0 & 96.0 & 98.0 & 98.0 & 94.0 & 96.0 & 98.0 & 62.0 & 92.8 \\
Qwen1.5-32B (Zh) & 94.0 & 96.0 & 94.0 & 92.0 & 92.0 & 90.0 & 94.0 & 82.0 & 91.8 \\
Qwen1.5-72B (En) & 100.0 & 94.0 & 98.0 & 94.0 & 96.0 & 94.0 & 96.0 & 92.0 & 95.5 \\
Qwen1.5-72B (Zh) & 94.0 & 90.0 & 92.0 & 86.0 & 88.0 & 94.0 & 98.0 & 96.0 & 92.2 \\
Qwen1.5-7B (En) & 96.0 & 92.0 & 94.0 & 92.0 & 96.0 & 90.0 & 98.0 & 88.0 & 93.2 \\
Qwen1.5-7B (Zh) & 94.0 & 96.0 & 94.0 & 90.0 & 88.0 & 94.0 & 94.0 & 98.0 & 93.5 \\
\bottomrule
\end{tabular}
\label{tab:model_accuracy_by_language}
\end{table*}


\begin{table*}[htbp]
\caption{Accuracy (\%) for English and Chinese with paraphrased data across different models and domains. Domain abbreviations: House=household routine, Nature=natural events, School=school life, Health=healthcare, Shop=shopping retail, Work=workplace activities, Trans=public transportation, Leisure=leisure recreation.}
\centering
\small
\begin{tabular}{lccccccccc}
\toprule
\textbf{Model} & \textbf{House} & \textbf{Nature} & \textbf{School} & \textbf{Health} & \textbf{Shop} & \textbf{Work} & \textbf{Trans} & \textbf{Leisure} & \textbf{Avg} \\
\midrule
Llama-3.2-1B (En) & 40.0 & 58.0 & 58.0 & 52.0 & 70.0 & 48.0 & 68.0 & 32.0 & 53.2 \\
Llama-3.2-1B (Zh) & 54.0 & 46.0 & 38.0 & 54.0 & 62.0 & 52.0 & 66.0 & 54.0 & 53.2 \\
Llama-3.2-3B (En) & 84.0 & 68.0 & 54.0 & 80.0 & 50.0 & 44.0 & 70.0 & 60.0 & 63.8 \\
Llama-3.2-3B (Zh) & 70.0 & 76.0 & 62.0 & 52.0 & 72.0 & 34.0 & 62.0 & 66.0 & 61.8 \\
Mistral-7B (En) & 90.0 & 98.0 & 90.0 & 96.0 & 98.0 & 86.0 & 94.0 & 98.0 & 93.8 \\
Mistral-7B (Zh) & 80.0 & 78.0 & 80.0 & 86.0 & 88.0 & 74.0 & 82.0 & 84.0 & 81.5 \\
Qwen1.5-1.8B (En) & 92.0 & 96.0 & 82.0 & 86.0 & 90.0 & 80.0 & 90.0 & 92.0 & 88.5 \\
Qwen1.5-1.8B (Zh) & 82.0 & 80.0 & 74.0 & 76.0 & 64.0 & 74.0 & 78.0 & 86.0 & 76.8 \\
Qwen1.5-14B (En) & 100.0 & 100.0 & 96.0 & 100.0 & 100.0 & 90.0 & 98.0 & 96.0 & 97.5 \\
Qwen1.5-14B (Zh) & 96.0 & 98.0 & 100.0 & 94.0 & 98.0 & 98.0 & 98.0 & 98.0 & 97.5 \\
Qwen1.5-32B (En) & 96.0 & 94.0 & 94.0 & 100.0 & 98.0 & 96.0 & 100.0 & 46.0 & 90.5 \\
Qwen1.5-32B (Zh) & 100.0 & 96.0 & 98.0 & 94.0 & 100.0 & 94.0 & 100.0 & 70.0 & 94.0 \\
Qwen1.5-72B (En) & 98.0 & 96.0 & 98.0 & 98.0 & 92.0 & 96.0 & 100.0 & 98.0 & 97.0 \\
Qwen1.5-72B (Zh) & 96.0 & 90.0 & 94.0 & 100.0 & 98.0 & 96.0 & 98.0 & 98.0 & 96.2 \\
Qwen1.5-7B (En) & 96.0 & 96.0 & 96.0 & 94.0 & 92.0 & 96.0 & 96.0 & 98.0 & 95.5 \\
Qwen1.5-7B (Zh) & 96.0 & 96.0 & 98.0 & 96.0 & 86.0 & 98.0 & 96.0 & 98.0 & 95.5 \\
\bottomrule
\label{tab:model_accuracy_by_para}
\end{tabular}
\end{table*}

\section{RCAR Distribution of Syntactic Components}\label{qwen_all}
\begin{figure*}[tb]
    \centering
    \includegraphics[width=1\linewidth]{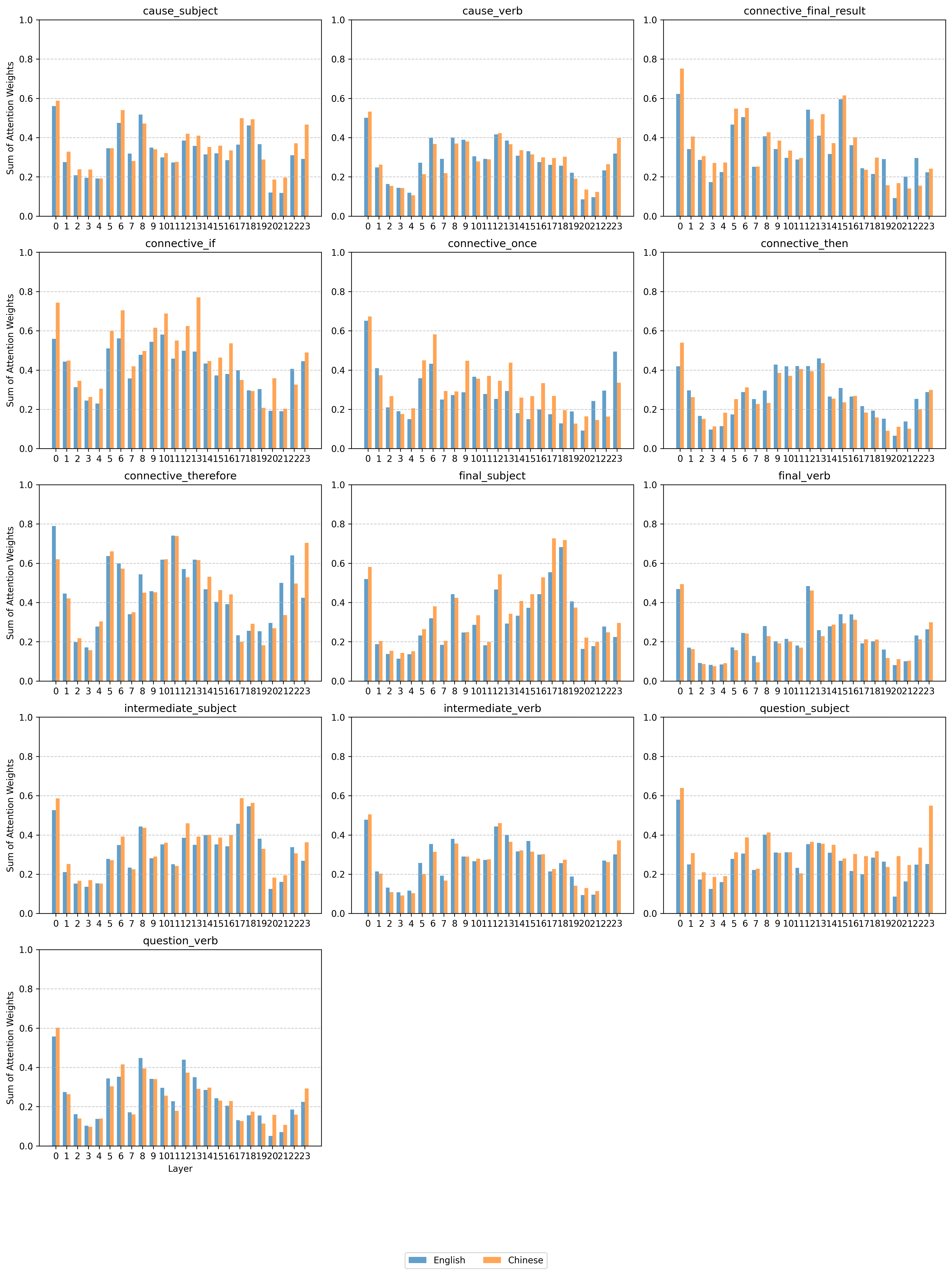}
    \caption{Layerwise RCAR over all syntactic components in Chinese (orange) and English (blue) causal chains. Despite similar layerwise trends, Chinese shows higher attention on subjects, while English focuses more on verbs.}
    \label{qwen_all_pic}
    
\end{figure*}

This section presents the layerwise RCAR distributions of all syntactic components in Qwen1.5–1.8B. See Figure~\ref{qwen_all_pic}.

\section{RCAR Distribution of Causal Components}\label{appendix_causal}
This section presents the RCAR distributions of causal components across different models.

\begin{figure*}[tb]
    \centering
    \includegraphics[width=0.8\linewidth]{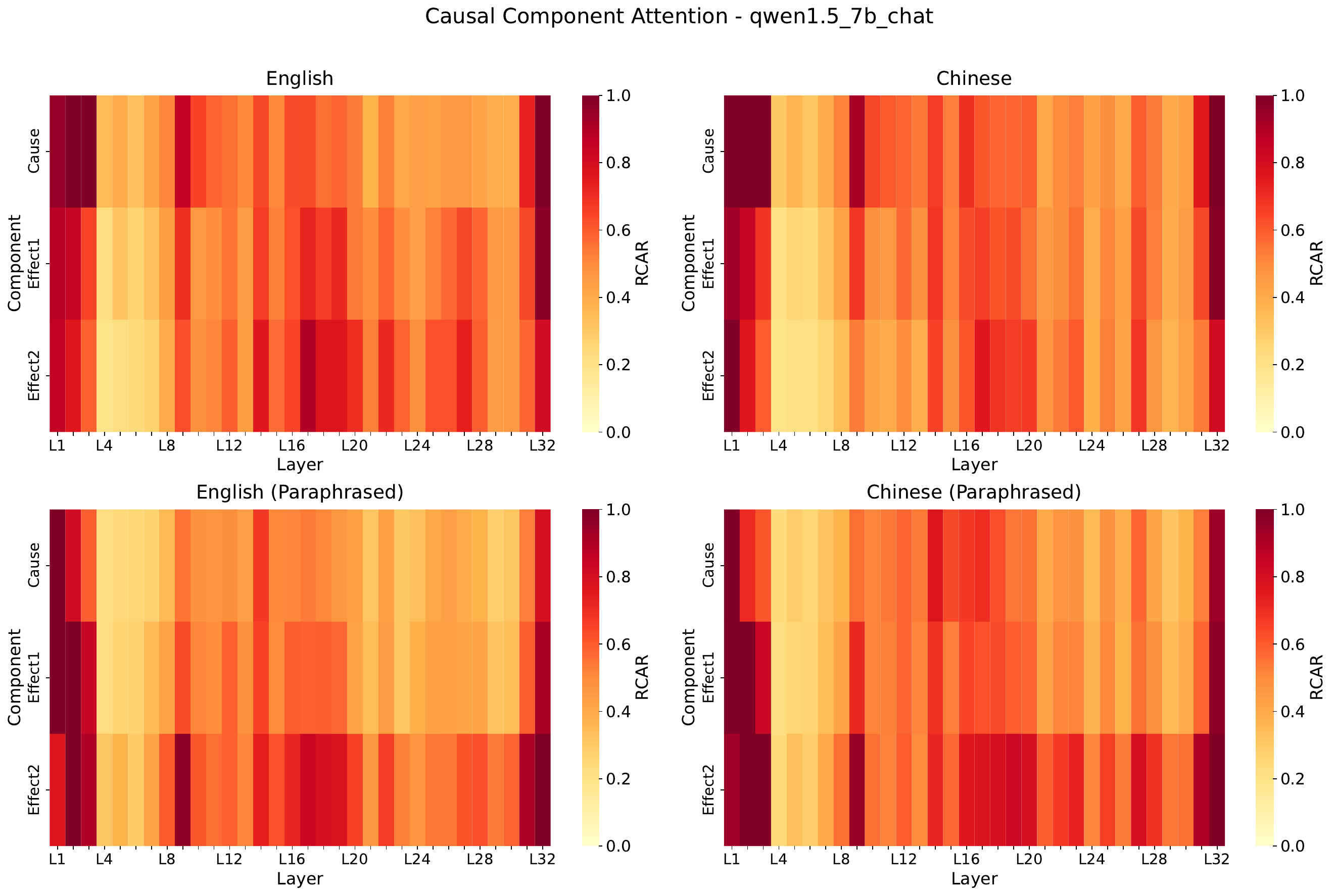}
    \caption{Layerwise RCAR heatmaps for Qwen1.5-7B-Chat showing attention patterns across causal components in four conditions. Note how attention distribution shifts significantly between forward and reversed causal chains, especially in Chinese.}
    \label{fig:qwen_7b_heatmap}
\end{figure*}

\begin{figure*}[tb]
    \centering
    \includegraphics[width=0.8\linewidth]{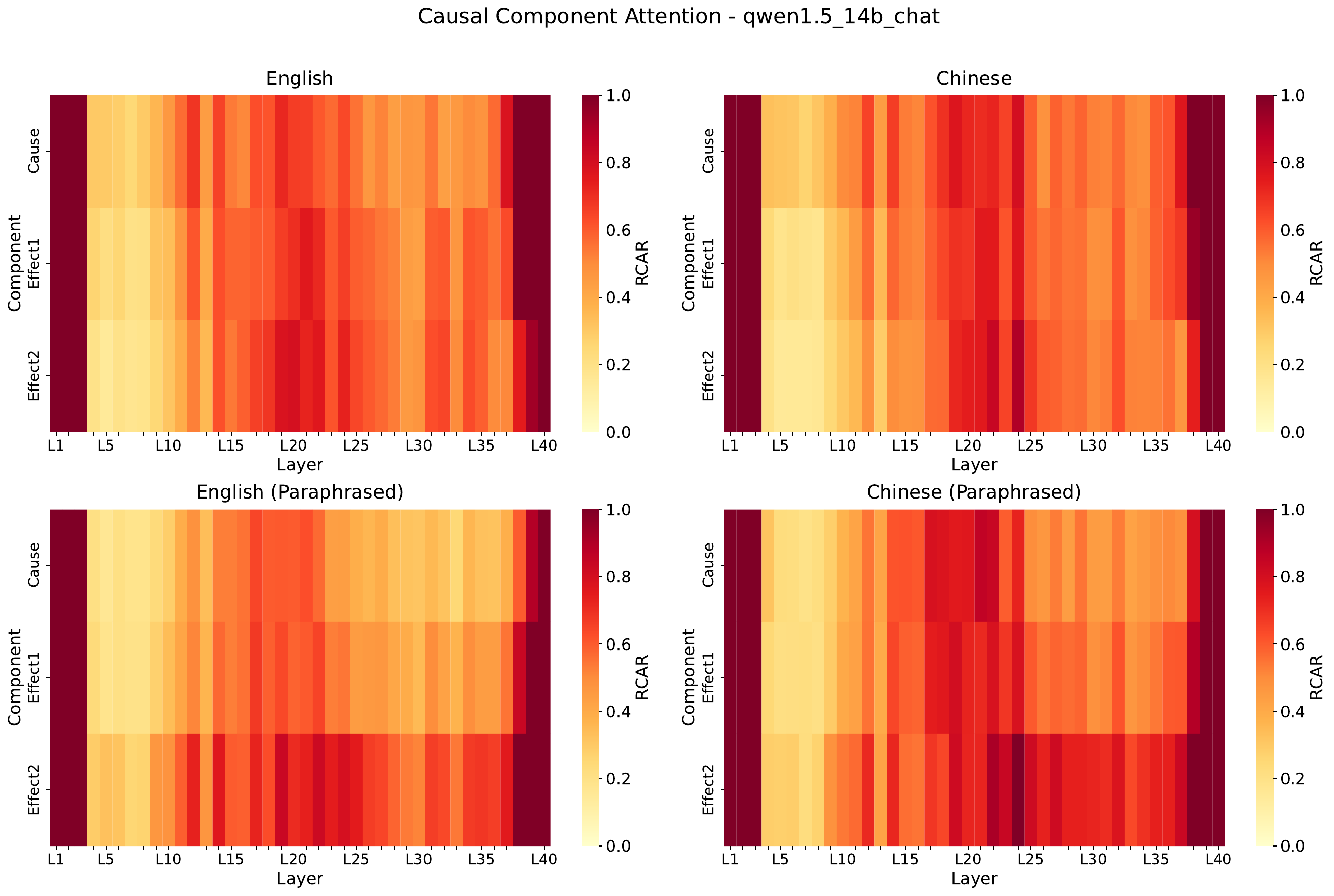}
    \caption{Layerwise RCAR heatmaps for Qwen1.5-14B-Chat. The larger model size shows more structured attention patterns compared to smaller variants, with clearer differentiation between causal components across languages.}
    \label{fig:qwen_14b_heatmap}
\end{figure*}

\begin{figure*}[tb]
    \centering
    \includegraphics[width=0.8\linewidth]{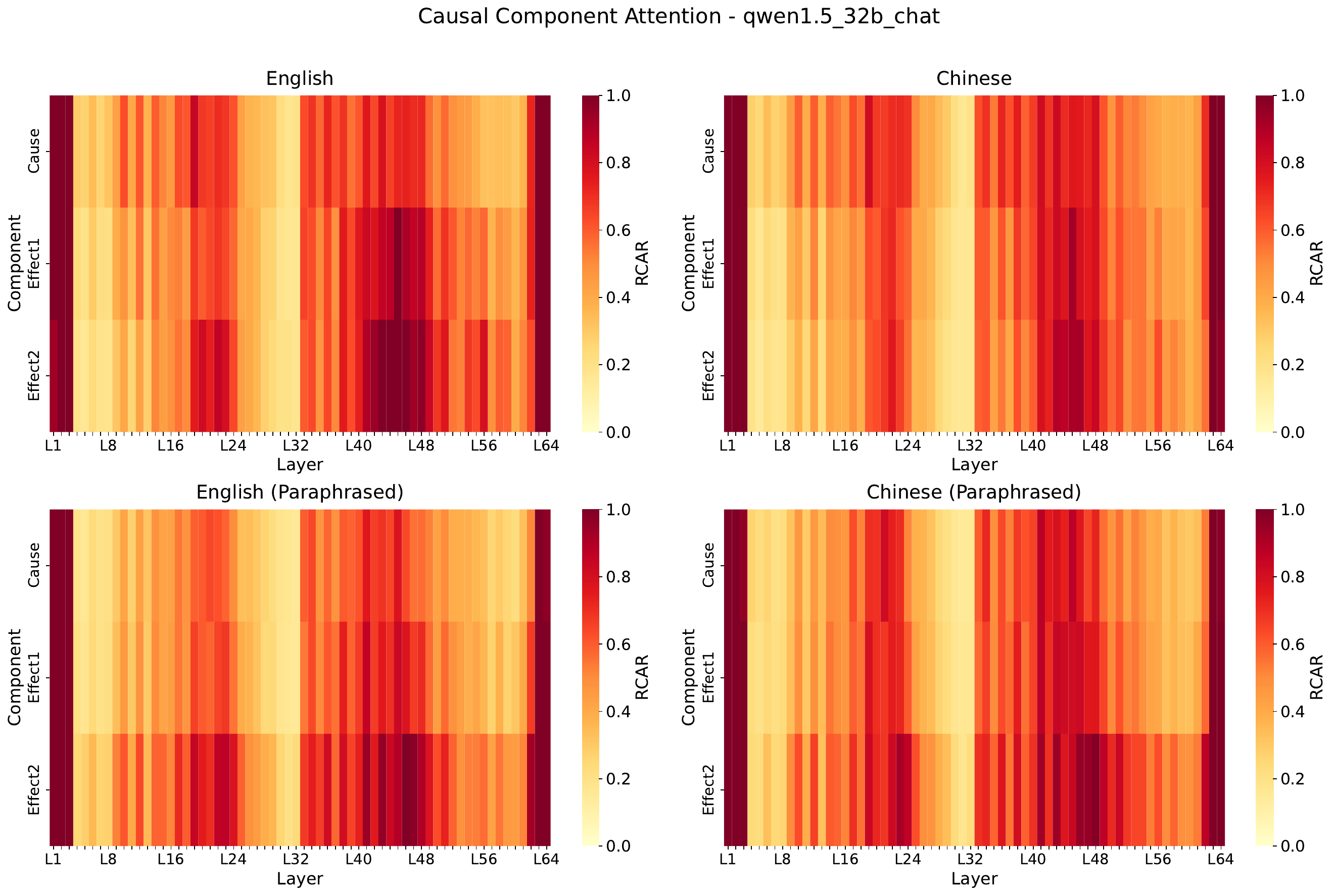}
    \caption{Layerwise RCAR heatmaps for Qwen1.5-32B-Chat. With 32 billion parameters, this model demonstrates more refined attention allocation strategies, with distinct patterns emerging in middle and later layers when processing causal information.}
    \label{fig:qwen_32b_heatmap}
\end{figure*}

\begin{figure*}[tb]
    \centering
    \includegraphics[width=0.8\linewidth]{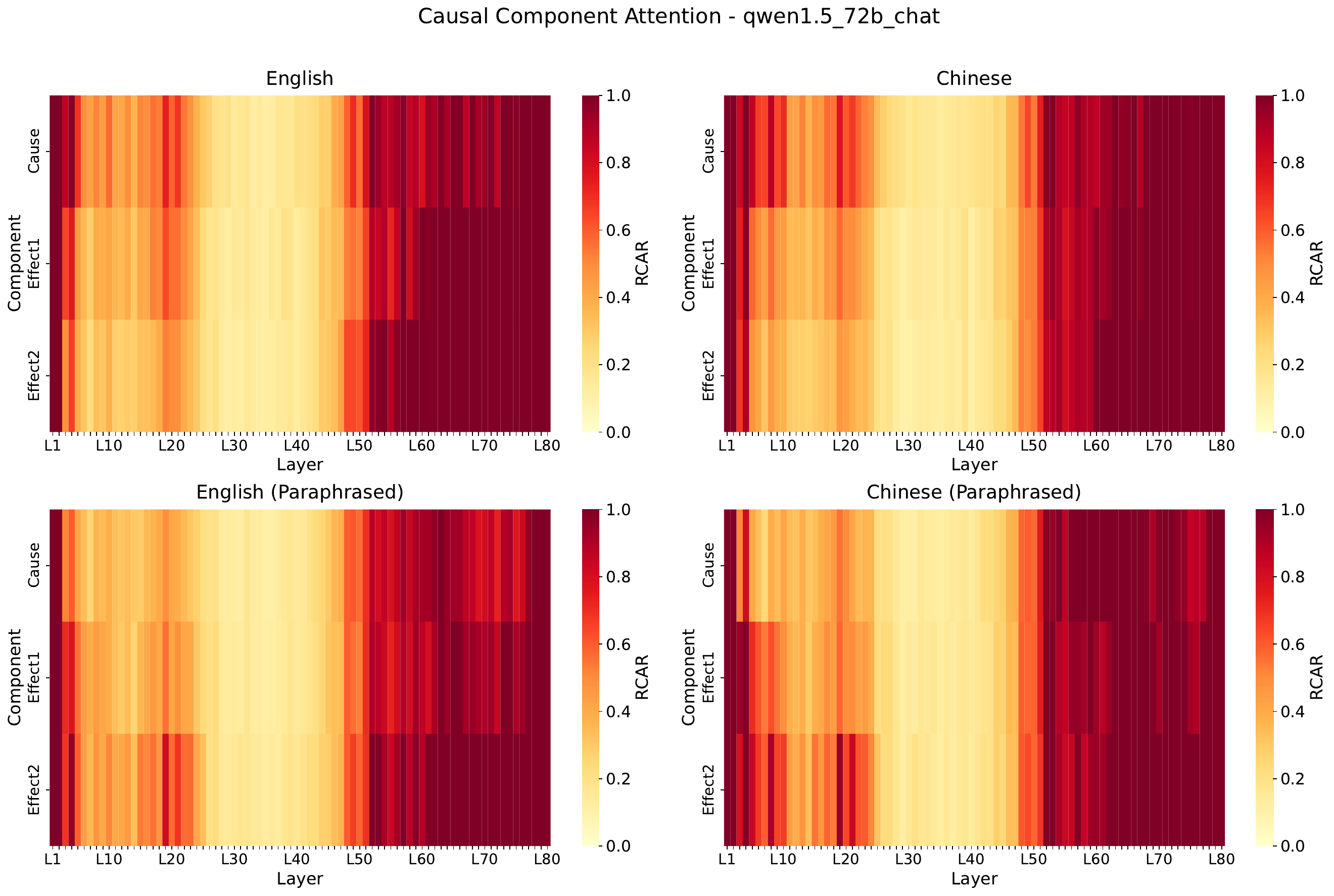}
    \caption{Layerwise RCAR heatmaps for Qwen1.5-72B-Chat. As the largest model in the Qwen family, it shows the most sophisticated attention patterns, with highly specialized layer functions and clearer cross-linguistic differences in causal processing.}
    \label{fig:qwen_72b_heatmap}
\end{figure*}

\noindent 
See Figures~\ref{fig:qwen_7b_heatmap},~\ref{fig:qwen_14b_heatmap},~\ref{fig:qwen_32b_heatmap}, and~\ref{fig:qwen_72b_heatmap}.

\end{document}